\pdfoutput=1

\documentclass[11pt]{article}

\usepackage[]{acl}

\usepackage{times}
\usepackage{latexsym}
\usepackage{amsmath,graphicx}
\usepackage{multirow,amsfonts,cite,array,url}
\usepackage{color}

\usepackage[T1]{fontenc}

\usepackage[utf8]{inputenc}

\usepackage{microtype}

\title{PALI-NLP at SemEval-2022 Task 4: Discriminative Fine-tuning of Transformers for Patronizing and Condescending Language Detection} 

\author{Dou Hu 
        \and Mengyuan Zhou 
        \and Xiyang Du
        \and Mengfei Yuan
        \\
        {\bf
        \and Meizhi Jin
        \and Lianxin Jiang
        \and Yang Mo
        \and Xiaofeng Shi
        } \\
        Ping An Life Insurance Company of China, Ltd. \\
        \texttt{\{HUDOU470, ZHOUMENGYUAN425, DUXIYANG037, YUANMENGFEI854, } \\
        \texttt{JINMEIZHI005, JIANGLIANXIN769, MOYANG853, SHIXIAOFENG309\}} \\
        \texttt{@pingan.com.cn} \\
}

\begin{document}
\maketitle
\begin{abstract}
Patronizing and condescending language (PCL) has a large harmful impact and is difficult to detect, both for human judges and existing NLP systems.
At SemEval-2022 Task 4, we propose a novel Transformer-based model and its ensembles to accurately understand such language context for PCL detection.
To facilitate comprehension of the subtle and subjective nature of PCL, two fine-tuning strategies are applied to capture discriminative features from diverse linguistic behaviour and categorical distribution. 
The system achieves remarkable results on the official ranking, including 1st in Subtask 1 and 5th in Subtask 2.
Extensive experiments on the task demonstrate the effectiveness of our system and its strategies.
\end{abstract}

\section{Introduction}
“\textit{Don’t worry, I know this is a mistake you usually make, we all make it sometimes, but I am bringing you a solution.}”, which is a typical example of Patronizing and Condescending Language (PCL) \citep{giles1993patronizing,huckin2002critical}, shows a superior attitude and apparent kindness towards others, while is generally expressed unconsciously. 
The impact of PCL can potentially be very harmful, as it feeds inequalities and routinizes discrimination \citep{ng2007language}, especially if it is geared towards vulnerable communities in the media. 
If we are able to detect and identify when we are condescending or patronizing towards others, a corrective action (e.g., a more inclusive message) could be taken for a more responsible communication. 

Recently, some works \citep{DBLP:conf/emnlp/WangP19,DBLP:conf/acl/SapGQJSC20} on PCL are gradually emerging in NLP community.  
Remarkably, \citet{perez2020don}
have shown that general pre-trained language models \citep{DBLP:conf/naacl/DevlinCLT19,DBLP:journals/corr/abs-1907-11692} can achieve nontrivial performance. 
However, the behaviour of PCL is usually more unconscious, subtler, and subjective than other harmful types of discourse that are widely studied, i.e., hate speech \citep{DBLP:conf/semeval/BasileBFNPPRS19}, offensive language \citep{DBLP:conf/semeval/ZampieriMNRFK19}, intended sarcasm \citep{duhjjms2022semeval}, fake news \citep{DBLP:conf/www/ZhangCLSZS21} and rumor \citep{DBLP:conf/acl/WeiHZYH20}. These characteristics make PCL detection a difficult challenge, both for human judges and existing NLP systems. 

To address this, we propose a novel Transformer-based model BERT-PCL (and its ensembles) with two \textbf{discriminative fine-tuning} strategies, to accurately understand such language context for PCL detection. 
The two strategies are grouped layer-wise learning rate decay (Grouped LLRD) and weighted random sampler (WRS), and both are beneficial for task-adaptive fine-tuning based on language models. 

A brief description of these two strategies is as follows:
\textbf{1)} As different layers capture different types of information \citep{yosinski2014transferable}, Grouped LLRD, a variant of LLRD \citep{DBLP:conf/acl/RuderH18,DBLP:conf/iclr/0007WKWA21}, is applied to group hidden layers of the pre-trained Transformer \citep{vaswani2017attention} into different sets and apply different learning rates to each in a certain extent. And then, we can make full use of different layers to capture more diverse and fine-grained linguistic features, which can boost understanding of the subtle and subjective nature of PCL. 
\textbf{2)} There is a quite common phenomenon that positive samples (patronizing or condescending) have a smaller number than the negative, which reflects usage rates of PCL in public forums \citep{DBLP:conf/emnlp/WangP19, perez2020don}. But the positive samples are more important when detecting PCL, due to the harmful impacts. To deal with imbalanced classes scenarios, we introduce WRS to place more emphasis on the minority classes. Under this strategy, our classifier can capture discriminative features from the categorical distribution and detect whether it contains PCL in an unbiased manner.


At SemEval-2022 Task 4 \citep{perezalmendros2022semeval}, our proposed system achieves \textbf{1st in Subtask 1} and \textbf{5th in Subtask 2} on the evaluation leaderboard\footnote{\url{https://sites.google.com/view/pcl-detection-semeval2022/ranking}}.
Meanwhile, in the post-evaluation phase, we further verified the results of the system on the test set of both subtasks. 
For Subtask 1, the single model BERT-PCL and its ensembles obtained \textbf{63.69\%} and \textbf{65.41\%} performance in terms of F1 of positive class, respectively.
For Subtask 2, the single model BERT-PCL and its ensembles obtained \textbf{43.28\%} and \textbf{45.66\%} performance in terms of macro-average F1, respectively.
Moreover, a series of experiments are conducted on the two subtasks of PCL detection. Results consistently demonstrate that our model and its ensembles significantly outperform comparison methods and the effectiveness of two strategies used in our system.

\section{Background}
\subsection{Task and Data Description}

The aim of SemEval-2022 Task 4 \citep{perezalmendros2022semeval} is to identify PCL, and to categorize the linguistic techniques (categories) used to express it, specifically when referring to vulnerable communities in the media.

This challenge is divided into two subtasks, each corresponding to a subset of the \textit{Don't Patronize Me!} (DPM) dataset \citep{perez2020don}.
The 10,469 annotated paragraphs (i.e., sentences in context) from the DPM corpus are used as training data, where each paragraph mentions one or several predefined vulnerable communities.
These paragraphs are collected using a keyword-based strategy and cover English language news sources from 20 different countries. 
A short description of the two subtasks and training data is as follows:
\begin{itemize}
    \item \textbf{Subtask 1: Binary classification}. Given a paragraph, a system must predict whether or not it contains any form of PCL. The training set consists of 10,469 paragraphs annotated with a label ranging from 0 to 4. Label 2, 3, and 4 means positive examples (condescending or patronizing) of PCL and the remaining labels means negatives. 
    \item \textbf{Subtask 2: Multi-label classification}. Given a paragraph, a system must identify the categories of PCL that are present.
    The 993 unique paragraphs (positive examples) in the training set, totaling 2,760 instances of PCL, are labeled with one or more PCL categories: 
    \textit{Unbalanced power relations}, \textit{Shallow solution}, \textit{Presupposition}, \textit{Authority voice}, \textit{Metaphor}, \textit{Compassion}, \textit{The poorer, the merrier}.
\end{itemize}
In addition, the test set for the evaluation phase contains around 4,000 manually annotated paragraphs with the PCL annotation scheme.
More details about the task can be found on the competition page\footnote{\url{https://competitions.codalab.org/competitions/34344}}. %


\subsection{Related Work}
Harmful language detection/recognition has been widely studied in various forms of discourse, such as hate speech \citep{DBLP:conf/semeval/BasileBFNPPRS19}, offensive language \citep{DBLP:conf/semeval/ZampieriMNRFK19}, intended sarcasm \citep{duhjjms2022semeval}, fake news \citep{DBLP:conf/www/ZhangCLSZS21} and rumors \citep{DBLP:conf/acl/WeiHZYH20,dou2021rumor}.  
Unlike these works generally focused on explicit, aggressive and flagrant phenomena, the study of patronizing and condescending Language (PCL) \citep{giles1993patronizing,huckin2002critical,chouliaraki2006spectatorship,margic2017communication} has been almost ignored in NLP community until recently.

To encourage more research on PCL language, \citet{DBLP:conf/emnlp/WangP19} present a condescension detection task and provides a \textit{TALKDOWN} dataset in comment-reply pairs from Reddit. 
Besides, \citet{perez2020don} introduce a \textit{Don’t Patronize Me!} dataset and the challenge of PCL detection towards vulnerable communities (e.g. refugees, homeless people, poor families).
These works establish several advanced baselines using pre-trained language models \citep{DBLP:conf/naacl/DevlinCLT19,DBLP:journals/corr/abs-1907-11692}, and suggest that detecting such language is a challenging task both for humans and NLP systems due to its subtle and subjective nature.


\section{System Overview}


In this section, we review our system adopted in SemEval-2022 Task 4, where we design a novel Transformer-based model BERT-PCL (and its ensembles) with two discriminative fine-tuning strategies for both subtasks of PCL detection. 

\subsection{Model Architecture}
BERT \citep{DBLP:conf/naacl/DevlinCLT19} uses masked language models to enable pretrained deep bidirectional representations, and can be fine-tuned to create task-specific models with powerful performance \citep{DBLP:conf/pkdd/WeiHZTZWHH20,DBLP:conf/seke/HuW20}.
Inspired by this, our system utilizes Transformers \citep{vaswani2017attention} to learn contextual representations of the input sentence under the BERT-like architecture.

Formally, given an input token sequence
$x_{i1}, ..., x_{iN}$ where $x_{ij}$ refers to $j$-th token in the $i$-th input sample, and $N$ is the maximum sequence length, the model learns to generate the context representation of the input token sequences: 
\begin{equation}
    \mathbf{h}_i = \text{BERT}(\text{[CLS]}, x_{i1}, ..., x_{iN}, \text{[SEP]}),
\end{equation}
where $\text{[CLS]}$ and $\text{[SEP]}$ are special tokens, usually at the beginning and end of each sequence, respectively.
$\mathbf{h}_i$ indicates the hidden representation of the $i$-th input sample, computed by the representation of [CLS] token in the last layer of the encoder. 

\subsection{PCL Detection}
\subsubsection{Subtask 1: Binary Classification}
Subtask 1, a binary classification task, aims to predict whether or not a paragraph contains any form of PCL.
After encoding, we apply a fully connected layer with the Softmax function to predict whether or not the input contains any form of PCL:
\begin{equation}
    \hat{\mathbf{y}}_i = Softmax(\mathbf{W} \mathbf{h}_i + \mathbf{b}),
\end{equation}
where $\mathbf{W}$ and $\mathbf{b}$ are trainable parameters.
We leverage Cross-entropy loss to optimize the system. The objective function of Subtask 1 is defined as:
\begin{equation}
    L = - \frac{1}{N}\sum_i (\mathbf{y}_i \log (\hat{\mathbf{y}}_i)  + (1- \mathbf{y}_i) \log (1-\hat{\mathbf{y}}_i))
\end{equation}
where $\mathbf{y}_i$ is the ground-truth label of PCL.

\subsubsection{Subtask 2: Multi-Label Classification}
Subtask 2 is a multi-label classification task. Its goal is to determine  which PCL categories a paragraph expresses. 
After encoding, we also apply a fully connected layer with the sigmoid function to predict the probability of each PCL class:
\begin{equation}
    \hat{\mathbf{y}}_i^c = \sigma(\mathbf{W}^c \mathbf{h}_{i} + \mathbf{b}^c),
\end{equation}
where $\sigma$ is the sigmoid function.
$\mathbf{W}^c$ and $\mathbf{b}^c$ are trainable parameters.
We use Binary Cross Entropy (BCE) loss \citep{DBLP:journals/pami/BengioCV13} for the multi-label classification task, denoted as: 
\begin{equation}
    L = - \frac{1}{N}\sum_i \sum_{c=1}^M [\mathbf{y}_{i}^c \log (\hat{\mathbf{y}}_i^c) + (1-\mathbf{y}_{i}^c) \log (1-\hat{\mathbf{y}}_i^c)  ],
\end{equation}
where $M$ is the number of classes, $\hat{\mathbf{y}}_i^c$ indicates the predicted probability that the $i$-th sample belongs to the $c$-th class.

\subsection{Fine-tuning Strategies}
For discriminative fine-tuning of the model, we introduce two strategies to boost the accurate understanding of PCL context, namely grouped layer-wise learning rate decay (Grouped LLRD) and weighted random sampler (WRS). 

\subsubsection{Grouped LLRD}
As different layers capture different types of information \citep{yosinski2014transferable},  they should be fine-tuned to different extents. Therefore, instead of using the same learning rate for all hidden layers of the Transformer, we tune each layer with different learning rates. 
Layer-wise learning rate decay (LLRD) \citep{DBLP:conf/acl/RuderH18,DBLP:conf/iclr/0007WKWA21} 
is a popular fine-tuning strategy that applies higher learning rates for top layers and lower learning rates for bottom layers. 
Inspired by this, we group layers into different sets and apply different learning rates to each, denoted as Grouped LLRD. 

Formally, we split all hidden layers of the Transformer into $G$ sets with embeddings attached to the first set.
The parameters of layers are denoted as $\{\theta^1,...,\theta^G\}$, where $\theta^g$ refers to the $g$-th group. 
The corresponding learning rate values are denoted as $\{ \eta^1, ...,\eta^G\}$, where $\eta^g$ indicates the learning rate of the $g$-th group. 
To capture discriminative features, a multiplicative decay rate $\lambda$ is used to change relative value of initial learning rates from adjacent groups in a controlled fashion. 
At time step $t$, the update of parameters $\theta$ is computed by:
\begin{equation}
    \theta^g_{t} = \theta^g_{t-1} - \eta^g \cdot \nabla_{\theta^g} J(\theta),
\end{equation}
where $\nabla_{\theta^g} J(\theta)$ is the gradient with regard to the model’s objective function.
The learning rate of the lower layer is applied as $\eta^{g-1} = \eta^g / \lambda$ during fine-tuning to decrease the learning rate group-by-group. 
In addition, same as LLRD, we use a learning rate that is slightly higher than the top hidden layer for the pooler head and classifier.

Under the above setting, we can capture more diverse and fine-grained linguistic features by flexibly optimizing different hidden layers of the Transformer.
It can boost understanding of PCL's subtle and subjective nature. 


\subsubsection{Weighted Random Sampler}
The PCL dataset is highly imbalanced, which causes problems for training the above models. 
To alleviate this imbalanced classes problem, we use a Weighted Random Sampler (WRS) to place more emphasis on the minority classes.
The samples are weighted and the probability of each sample being selected is determined by its relative weight.

For both subtasks, the sampling weight of the $i$-th sample is computed by: 
\begin{equation}
    s_i = \left\{
\begin{aligned}
1/\sqrt{\kappa_p}, & \ \  \text{if it contains PCL,} \\
1/\sqrt{\kappa_n}, & \ \  \text{otherwise,}
\end{aligned}
\right.
\end{equation}
where $\kappa_{p}$ and $\kappa_{n}$ refer to the ratio of positive and negative examples of PCL in training data, respectively. 
Then, the elements are sampled based on the passed weights. 
It is worth noting that the number of samples is equal to the length of the training set. 
During training, the sampler tends to select samples from positive examples with small data volume. 
In this way, we can have positive and negative classes with equal probability. And the classifier can capture discriminative features from categorical distribution in an unbiased manner.


\subsection{Ensemble}
For the final submissions, we apply a voter-based fusion technique \citep{DBLP:conf/sspr/MorvantHA14} to ensemble several BERT-PCL models.
Concretely, we train the proposed BERT-PCL with five different random seeds. 
Then, we select Top-3 models according to average result of k-fold cross-validation on the training data.
Finally, results of the test set predicted by the three optimal models are voted to get the final submission.

\section{Experimental Setup}
\subsection{Comparison Methods}

We compare BERT-PCL and its ensembles with the following several methods:
\begin{itemize}
    \item \textbf{Random} is based on random guessing, choosing each class/label with an equal probability.
    \item \textbf{BERT}  \citep{DBLP:conf/naacl/DevlinCLT19} is a language model pre-trained in a self-supervised fashion based on deep bidirectional transformers.  We use \textit{bert-base-uncased}\footnote{\url{https://huggingface.co/}\label{code}} to initialize BERT. 
    \item \textbf{ALBERT} \citep{DBLP:conf/iclr/LanCGGSS20} presents two parameter-reduction techniques to lower memory consumption and increase the training speed of BERT. 
    We use \textit{albert-large-v2}\textsuperscript{\ref{code}} to initialize ALBERT. 
    \item \textbf{ERNIE 2.0} \citep{DBLP:conf/aaai/SunWLFTWW20} is a continual pre-training framework, which builds and learns incrementally pre-training tasks through constant multi-task learning.
    We use \textit{nghuyong/ernie-2.0-large-en}\textsuperscript{\ref{code}} to initialize ERNIE 2.0. 
    \item \textbf{RoBERTa} \citep{DBLP:journals/corr/abs-1907-11692} optimizes the training procedure of BERT and removes the next sentence predict objective when pre-training.
    We use \textit{roberta-large}\textsuperscript{\ref{code}} to initialize RoBERTa. 
\end{itemize}

\subsection{Implementation Details}
For both subtasks, stratified k-fold cross validation  \citep{DBLP:conf/ijcai/Kohavi95,DBLP:conf/pkdd/SechidisTV11} is performed to split limited training data into 5 folds.
We choose the optimal hyperparameter values based on the the average result of validation sets for all folds, and evaluate the performance of systems on the test data.
BERT-PCL is initialized with the \textit{roberta-large}\textsuperscript{\ref{code}} parameters, due to the nontrivial and consistent performance in both subtasks. 
Following \citet{perez2020don,perezalmendros2022semeval}, the evaluation metrics are F1 over the positive class for Subtask 1 and macro-average F1 for Subtask 2.

We group layers into $3$ groups, i.e., $G=3$. 
The learning rate for layers in the lower, median, and higher groups as $\eta / \lambda$, $\eta$, and $\eta * \lambda$, respectively, where $\eta$ is set to 1e-5. $\lambda$ is a hyperparameter searched from $\{0.6, 1.6, 2.6, 3.6, 4.6, 5.6, 6.6\}$. 
For the training of BERT-PCL, the optimal value of $lambda$ is 1.6 for Subtask 1 and 3.6 for Subtask 2.
The experiments are conducted with batch size of 4, maximum length of 250, and dropout rate of 0.4. 
The number of epochs is set to 10 and the maximum patience number of early stopping is set to 50 batches. 
AdamW optimizer \citep{DBLP:journals/corr/KingmaB14} is used with a weight decay of 0.01.
A cosine annealing schedule \citep{DBLP:conf/iclr/LoshchilovH17} is applied to decay the learning rate, with a linear warmup for the first 10\% steps.

To effectively utilize the country term and search keyword term corresponding to each paragraph in the corpus, we concatenate these terms with the original paragraph as the input sequence. In the implementation, two special token pairs (i.e., <e> and </e>) are introduced as the term boundary.

\begin{table}[t]
\centering
\resizebox{0.99\linewidth}{!}{
\begin{tabular}{l|ccc}
\hline
\multicolumn{1}{c|}{\multirow{1}{*}{\textbf{}}} & {P} & {R} & {F1}  \\
\hline
Random      & 8.98	& 55.21	& 15.45 \\ 
BERT        & 56.20	& 48.58	& 52.12 \\
ALBERT      & 59.43 & 32.81	& 42.28 \\
ERNIE 2.0   & 59.24	& 58.68	& 58.95 \\
RoBERTa     & 60.65	& 64.67	& 62.60 \\
\textbf{BERT-PCL}    & 64.31	& 63.09	&  \textbf{63.69} \\ 
\hline
Ensemble 1.0 $^\dagger$              & 64.60	& 65.62	&  65.10 \\ 
\textbf{Ensemble 2.0}     & 65.20	& 65.62	&  \textbf{65.41} \\ 
\hline
\end{tabular}
}
\caption{
Results for the problem of detecting PCL, viewed as a binary classification problem (Subtask 1). 
The results are reported in terms of the precision (P), recall (R) and F1 score of the positive class. 
All compared pre-trained models are fine-tuned on the task dataset.
$^\dagger$ indicates the results on the official ranking.
}
\label{tab:result_1}
\end{table}

\begin{table*}[t]
\centering
\resizebox{1.0\linewidth}{!}{
\begin{tabular}{l|cccccc|cc}
\hline
\multicolumn{1}{c|}{\multirow{1}{*}{}} 
& Random 
& BERT 
& ALBERT
& ERNIE 2.0
& RoBERTa 
& \textbf{BERT-PCL} 
& Ensemble 1.0 $^\dagger$
& \textbf{Ensemble 2.0}  \\   
\hline
\textit{unb.}    & 10.82 & 51.52 & 51.31 & 56.50 & 57.41 & 58.90 & 62.78 & 61.40 \\ 
\textit{shal.}      & 2.23  & 40.62 & 41.79 & 55.88 & 50.60 & 50.55 & 54.76 & 55.81 \\ 
\textit{pres.}     & 3.03  & 20.29 & 16.39 & 25.97 & 26.83 & 42.86 & 34.15 & 39.13 \\ 
\textit{auth.}   & 4.21  &  6.98 &  9.64 & 16.49 & 26.79 & 28.07 & 34.19 & 34.15 \\
\textit{met.}           & 1.91  &  8.70 &  9.52 & 28.07 & 40.51 & 40.00  & 33.33 & 43.84 \\
\textit{comp.}         & 5.98  & 42.48 & 40.00 & 45.90 & 48.28 & 49.24 & 50.82 & 49.61 \\ 
\textit{merr.}   & 1.22  &  0.00 &  0.00 &  0.00 & 14.81 & 33.33 &  8.70 & 35.71 \\ 
\hline
{{Average}}                 & 4.20  & 24.37 & 24.09 & 32.69 & 37.89 & \textbf{43.28} & 39.82 & \textbf{45.66} \\
\hline
\end{tabular}
}
\caption{
Results for the problem of categorizing PCL, viewed as a paragraph-level multi-label classification problem (Subtask 2).
We report the per-class F1 and macro-average F1. 
All compared pre-trained model are fine-tuned on the task dataset.
$^\dagger$ indicates the results on the official ranking.
The considered seven categories are 
\textit{Unbalanced power relations} (unb.), \textit{Shallow solution} (shal.),
\textit{Presupposition} (pres.), \textit{Authority voice} (auth.),  
\textit{Metaphor} (met.), \textit{Compassion} (comp.)
and \textit{The poorer, the merrier} (merr.).
}
\label{tab:result_2}
\end{table*}

\section{Result and Discussion}
\subsection{Overall Result}

The overall results in both subtasks are summarized in 
Table~\ref{tab:result_1} and \ref{tab:result_2}.
Unsurprisingly, all pre-trained models clearly outperform the random baseline. 
The proposed BERT-PCL and its ensembles (i.e., Ensemble 1.0 and Ensemble 2.0) consistently obtain the best performance than the comparison methods on both subtasks.
Specifically, BERT-PCL gains 1.09\% and 5.39\% absolute improvements for Subtask 1 and 2, respectively.
These results show the superiority of our models.

In Table~\ref{tab:result_1}, both Ensemble 1.0 and Ensemble 2.0 are fused by three
optimal full BERT-PCL models with different seeds and obtain a better performance than BERT-PCL in Subtask 1.
In Table~\ref{tab:result_2}, Ensemble 1.0 is fused by the three optimal BERT-PCL that removes WRS, since we found that it is worse when performing WRS according to weights of category label in Subtask 2. Different from it, in the post-evaluation phase, we perform WRS according to weights of positive samples (patronizing or condescending) and fuse three optimal full BERT-PCL as Ensemble 2.0. As shown in Table~\ref{tab:result_2}, Ensemble 2.0 obtains a better performance than BERT-PCL and Ensemble 1.0 in Subtask 2.

\begin{figure}[t]
    \centering
    \includegraphics[width=\linewidth]{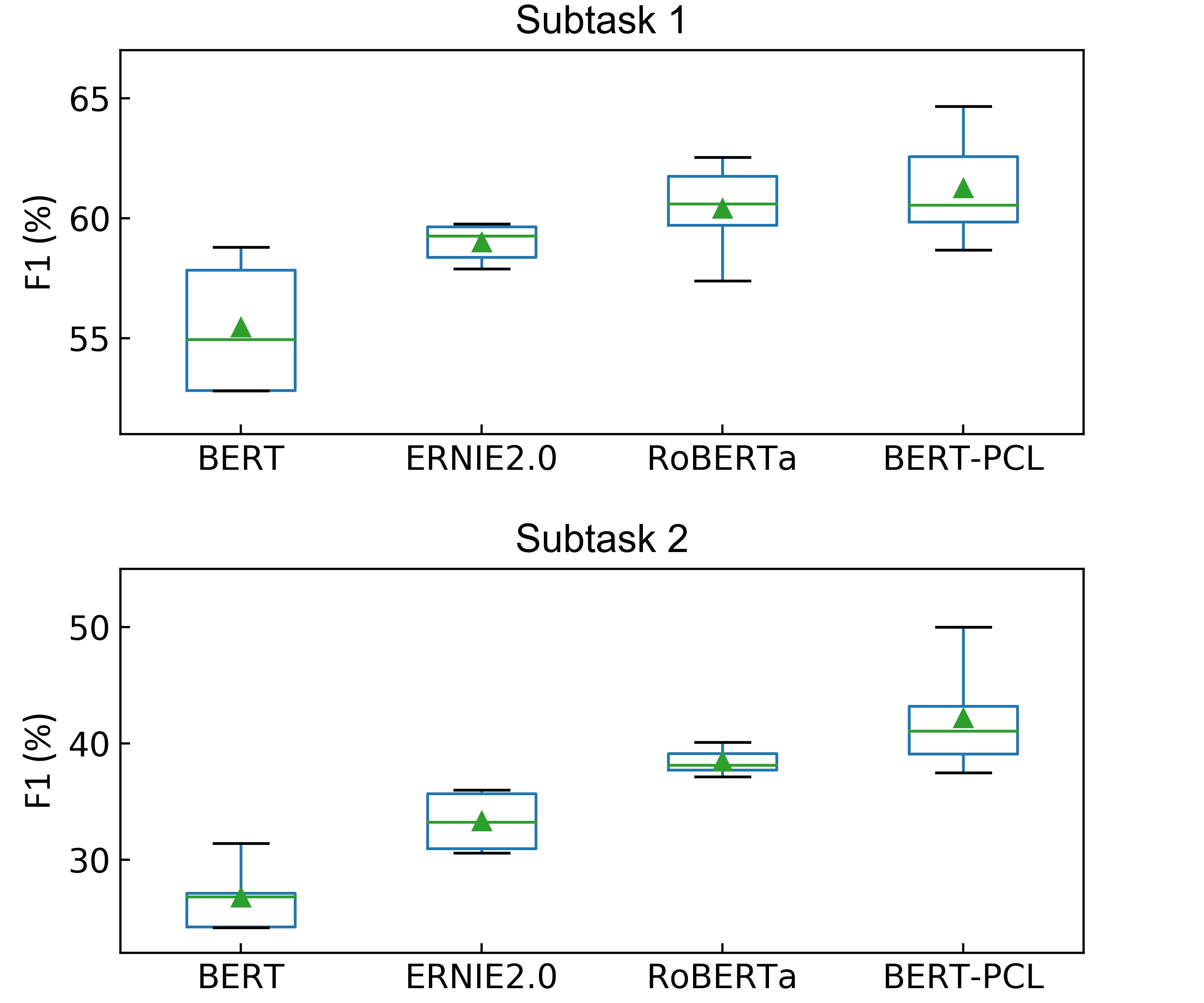}
    \caption{Results on the validation set for both subtasks. 
    The box displays the distribution of results where the green triangle indicates the mean of results, the green line and two blue lines represent the 25\%, 50\%, and 75\% quartiles, respectively, and black lines are the maximum and minimum values.
    For Subtask 1, we report F1 score of the positive class; and for Subtask 2, we list macro-average F1 score.
    }
    \label{fig:dev_kFold}
\end{figure}

Then, we qualitatively analyze the performance of BERT-PCL and typical baselines on the validation set for both subtasks.
The results are illustrated in Figure~\ref{fig:dev_kFold}.
From the figure, BERT-PCL consistently obtains the best performance on the validation set for both subtasks, which confirms again the superiority of the proposed method.

\subsection{Ablation Study}

\begin{figure}[t]
    \centering
    \includegraphics[width=\linewidth]{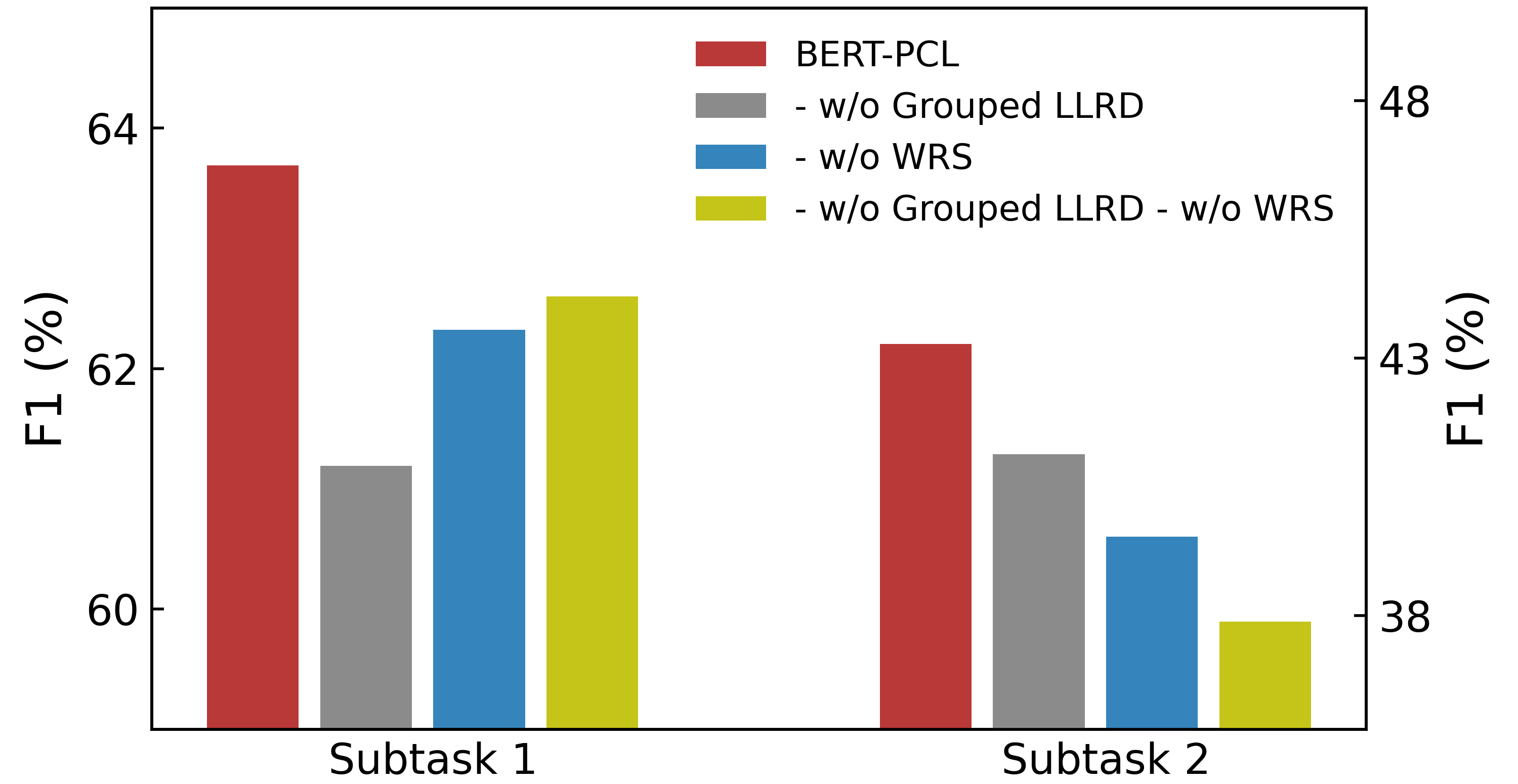}
    \caption{Results of ablation study for PCL detection.
    For Subtask 1, we report F1 score of the positive class; and for Subtask 2, we list macro-average F1 score.
    }
    \label{fig:result_abla}
\end{figure}

We conduct ablation studies by removing key components of BERT-PCL:  
1) \textbf{- w/o Grouped LLRD} refers to removing the Grouped LLRD.
2) \textbf{- w/o WRS} refers to removing the WRS.
3) \textbf{- w/o Grouped LLRD - w/o WRS} refers to removing both Grouped LLRD and WRS, degenerated to RoBERTa.

Figure~\ref{fig:result_abla} shows results of ablation studies on two subtasks of PCL detection.
The full model yields the best performance on both subtasks. 
When removing either Grouped LLRD or WRS, the results of variants decline significantly on both subtasks. 
Specifically, when only removing Grouped LLRD, the model achieves 2.50\%  and 2.14\% degradation of performance in Subtask 1 and 2, respectively.
When only removing WRS, the results decline by 1.37\% and 3.74\% in terms of F1 scores in Subtask 1 and 2, respectively.
The above results consistently indicate the effectiveness of the two components.

When removing both components, the performance also decreases on both subtasks.
Note that \textbf{- w/o Grouped LLRD - w/o WRS} achieves a better F1 score of the positive class than \textbf{- w/o WRS} or \textbf{- w/o Grouped LLRD} in Subtask 1.
This can be explained that ignoring Grouped LLRD limits to explore diverse features of positive samples, and further removing WRS may magnify this limitation due to the imbalanced class problem.
Therefore, the model with two modules removed yields slightly better results than ablation models with only one module removed.
Different from Subtask 1, Subtask 2 is a multi-label classification problem and we report the macro-average F1 score.
Using Grouped LLRD can capture diverse features of each category label in positive samples, and WRS according to weights of positive and negative samples further promotes the model's attention to positive samples.
Hence, removing both modules obtains the worst performance in Subtask 2.

\subsection{Parameter Analysis}

\begin{figure}[t]
    \centering
    \includegraphics[width=\linewidth]{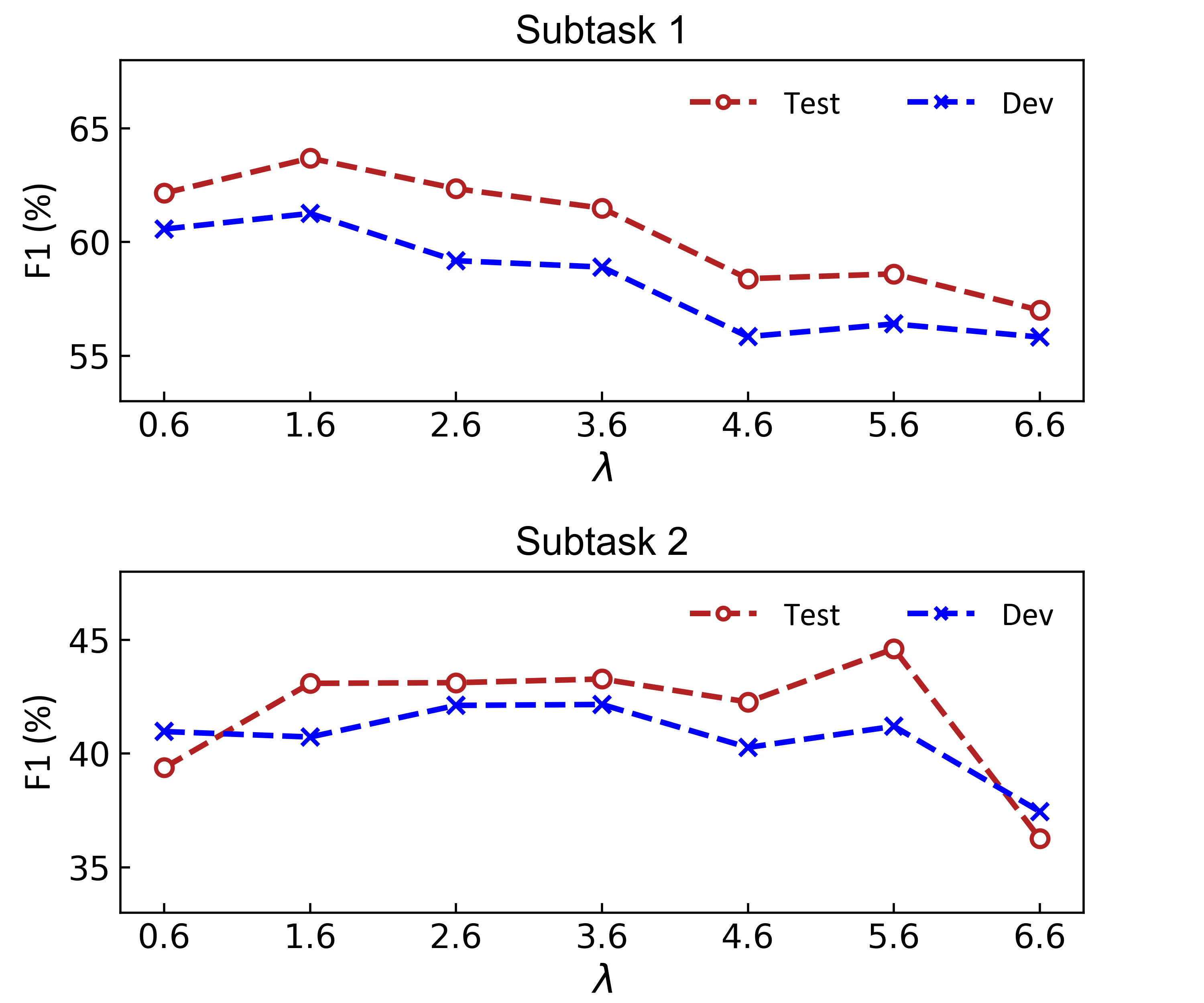}
    \caption{Results against different values of hyperparameter $\lambda$ in Grouped LLRD on both test and validation sets.
    We report F1 score of the positive class for Subtask 1, and list macro-average F1 score for Subtask 2.}
    \label{fig:llrd}
\end{figure}

In this part, we explore the performance of BERT-PCL against different $\lambda$ in Grouped LLRD.  
Bottom groups often encode more general and broad-based information, while top groups closer to the output encode information more localized and specific to the task on hand. In our model, a suitable value $\lambda$ can control and balance these different layers of the Transformer to capture different kinds of features from diverse linguistic behaviour. 

The results are illustrated in Figure~\ref{fig:llrd}.
Based on k-fold cross validation on training data, we find the local optimum of $\lambda$, 1.6 for Subtask 1 and 3.6 for Subtask 2, and the resulting model consistently performed excellently on the test set. 
Under the optimal setting, different layers in the model can capture more diverse and fine-grained linguistic features, enhancing the understanding of the subtle and subjective nature of PCL.
However, larger $\lambda$ would make the model overfit to small datasets and suffer catastrophic forgetting during fine-tuning. 
Hence, the performance degrades as $\lambda$ increases.

\begin{table*}[t]
  \centering
  \resizebox{1.0\linewidth}{!}{
  \begin{tabular}{cp{9.5cm}cccc} 
  \hline
  \multirow{2}{*}{No.} & \multirow{2}{*}{{Para.}}  
  & \multirow{2}{*}{Gold} & Pred.  & Pred.  & Pred.  \\
  & & & (BERT-PCL) & (RoBERTa) & (ERNIE 2.0) \\  \hline 
  1
  & \textit{“Jesus is the Master Feminist because he championed the cause of women,” she said.}
  &  pos. & pos. & neg. & neg. \\  
  2
  & \textit{There is a saying that goes "A friend in need is a friend indeed. " This means, a good friend is the one who rescues a friend trapped in unsolved problems.}
  &  neg. & neg. & neg. & pos. \\  
  3
  & \textit{"The government is implementing several schemes that would change the economic position of poor families," she added.}
  &  pos. & neg. & neg. & neg. \\  
  4
  & \textit{Alexis and her family decided to donate more than 400 of those presents to children in need.}
  &  neg. & pos. & pos. & pos. \\  
  \hline
  \end{tabular}
  }
  \caption{
  Case studies in Subtask 1: Binary Classification. The table shows four examples of paragraphs, their gold labels and predictions by three methods (BERT-PCL, RoBERTa and ERNIE 2.0). 
  The pos. means the positive class of PCL, i.e. as instances containing PCL. 
  Likewise, the neg. means the negatives. 
  }
\label{tab:case}
\end{table*}

\begin{table*}[t]
  \centering
  \resizebox{1.0\linewidth}{!}{
  \begin{tabular}{cp{8.5cm}cccc} 
  \hline
  \multirow{2}{*}{No.} & \multirow{2}{*}{{Para.}}   
  & \multicolumn{1}{c}{\multirow{2}{*}{Gold}} & \multicolumn{1}{c}{Pred.}  
  & \multicolumn{1}{c}{Pred.}  
  & \multicolumn{1}{c}{Pred.}  \\
  & & & {(BERT-PCL)}   &  {(RoBERTa)}    &  {(ERNIE 2.0)} \\  \hline 
  1
  & \textit{Through Gawad Kalinga, Meloto has proven to be a key player in the housing industry, helping provide decent homes and sustainable livelihood to the marginalized and homeless Filipinos.}
  &  unb., comp.         & unb., comp.    & -             & unb., comp. \\  
  2
  & \textit{Pope Francis will visit a tiny Italian island to greet refugees and immigrants, pray for those who have lost their lives at sea and call for greater solidarity.}
  &  unb., shal.         & unb., shal.    & -             & - \\  
  3
  & \textit{In South Africa, education is a right and not a privilege, but an unfavourable background can unconsciously infringe on this right.}
  &  unb., pres.,  met.  &  unb., met.   & unb., auth.   & unb. \\  
  4
  & \textit{Thankfully, while Krishna Tulasi can’t entirely escape from the trope of disabled persons with hearts of gold, it manages to do better than many previous films with disabled protagonists.}
  &  merr. & - & - & - \\  
  \hline
  \end{tabular}
  }
  \caption{
  Case studies in Subtask 2: Multi-Label Classification. 
  The table shows four examples of paragraphs, their gold labels and predictions by three methods (BERT-PCL, RoBERTa and ERNIE 2.0).
  The categories stand for:
  \textit{Unbalanced power relations} (unb.), \textit{Shallow solution} (shal.),
  \textit{Presupposition} (pres.), \textit{Authority voice} (auth.),  
  \textit{Metaphor} (met.), \textit{Compassion} (comp.)
  and \textit{The poorer, the merrier} (merr.).
  }
\label{tab:case2}
\end{table*}

It is worth noting that in when $\lambda$ is up to 5.6 for Subtask 2, the model achieves suboptimal results on the validation set but performs exceptionally well on the test set.
This may be because the model overfits some redundant features of the corpus.
\subsection{Case Study}

Table~\ref{tab:case} and Table~\ref{tab:case2} show several typical examples in the training set from Subtask 1 and 2, respectively. Their gold labels and predictions by BERT-PCL, RoBERTa and ERNIE 2.0 are presented in the corresponding columns. 

In Table~\ref{tab:case},  the first case is correctly classified by BERT-PCL, while is misclassified by other methods. We can easily observe that this example has the characteristics of \textit{Unbalanced power relations} and \textit{Authority voice}, and the language expression of the latter is more subtle. Unlike other methods, BERT-PCL can capture  the linguistic phenomena of PCL through a discriminative fine-tuning process, and thus detect them correctly.
For the second, BERT-PCL and RoBERTa can accurately identify the positive paragraphs, using the sentence representation ability learned by the pre-trained model. 
The latter two examples are consistently predicted as false negatives and false positives by all methods, respectively. We notice that both paragraphs have been annotated by two human annotators as borderline PCL. Unsurprisingly, these methods also struggle to detect such cases. 

As seen in Table~\ref{tab:case2}, only BERT-PCL can correctly determine fine-grained PCL categories of the first two cases, which again illustrates the superiority of our method.
It can be noticed that the third example has three PCL sub-categories (i.e., unb., pres., met.) with a certain internal correlation, and the gold label (i.e., merr.) of the fourth example appears too little in the training set. These phenomena increase the difficulty of identifying the two examples, which leads to wrong predicted labels. We believe that identifying multiple related sub-categories simultaneously and controlling the imbalance of positive PCL labels are urgent challenges for Subtask 2.



\section{Conclusion}
In this paper, we propose an advanced BERT-like model and its ensembles to accurately understand and detect patronizing and condescending language. 
Based on the pre-trained Transformer, we apply two fine-tuning strategies to capture discriminative features from diverse linguistic behaviour and categorical distribution. 
At SemEval-2022 Task 4, our system achieves 1st in Subtask 1 and 5th in Subtask 2 on the official ranking. 
Extensive experiments demonstrate the effectiveness and superiority of the proposed system and its strategies.

\section*{Acknowledgements}
The research is supported by the Ping An Life Insurance. 
All the work in this paper are conducted during the SemEval-2022 Competition.
Last but not least, we thank the SemEval-2022 organizers for their effort in preparing the challenge, and the reviewers for their insightful and constructive comments. 

\end{document}